\documentclass[10pt,twocolumn,letterpaper]{article}

\usepackage{cvpr}      

\usepackage{graphicx}
\usepackage{amsmath}
\usepackage{amssymb}
\usepackage{booktabs}
\usepackage{multirow}
\usepackage{verbatim}
\usepackage{threeparttable}
\usepackage{color}

\definecolor{hao}{rgb}{0.0,0.5,0.0}

\definecolor{qz}{rgb}{1.0,0.0,0.0}

\definecolor{todo}{rgb}{0.0,0.0,1.0}

\definecolor{zjy}{rgb}{0.0,0.0,1.0}

\newcommand{\ours}{TSCODE}
\newcommand{\ptitle}[1]{\noindent\textbf{#1}}

\newcommand{\tight}[1]{\hspace{2pt}{#1}{\hspace{2pt}}}

\usepackage[pagebackref,breaklinks,colorlinks]{hyperref}

\usepackage[capitalize]{cleveref}
\crefname{section}{Sec.}{Secs.}
\Crefname{section}{Section}{Sections}
\Crefname{table}{Table}{Tables}
\crefname{table}{Tab.}{Tabs.}

\def\cvprPaperID{2704} 
\def\confName{CVPR}
\def\confYear{2023}

\begin{document}

\title{Task-Specific Context Decoupling for Object Detection}

\author{
Jiayuan Zhuang$^1$\hspace{10pt}
Zheng Qin$^1$\hspace{10pt}
Hao Yu$^2$\hspace{10pt}
Xucan Chen$^1$\\
$^1$\text{National University of Defense Technology}\hspace{10pt} $^2$\text{Technical University of Munich}\\
\small\texttt{
alpc111@163.com\hspace{5pt}
qinzheng12@nudt.edu.cn\hspace{5pt}
hao.yu@tum.de\hspace{5pt}
xcchen18@139.com
}
}

\maketitle


\begin{abstract}
Classification and localization are two main sub-tasks in object detection. Nonetheless, these two tasks have inconsistent preferences for feature context, \ie, localization expects more boundary-aware features to accurately regress the bounding box, while more semantic context is preferred for object classification. Exsiting methods usually leverage disentangled heads to learn different feature context for each task. However, the heads are still applied on the same input features, which leads to an imperfect balance between classifcation and localization. In this work, we propose a novel Task-Specific COntext DEcoupling (\emph{TSCODE}) head which further disentangles the feature encoding for two tasks. For classification, we generate spatially-coarse but semantically-strong feature encoding. For localization, we provide high-resolution feature map containing more edge information to better regress object boundaries. TSCODE is plug-and-play and can be easily incorperated into existing detection pipelines. Extensive experiments demonstrate that our method stably improves different detectors by over 1.0 AP with less computational cost. Our code and models will be publicly released.
\end{abstract}
\vspace{-10pt}


\section{Introduction}
\label{sec:intro}

Object detection aims to recognize and localize objects existing in a natural image, which is a fundamental but challenging task in many computer vision applications. Recent advances in object detection have been predominated by deep learning-based methods~\cite{ren2015faster,he2015spatial,redmon2016you,girshick2014rich,lin2017focal}, where the task is typically formulated as the \emph{classification} of regions of interest and the \emph{localization} of bounding boxes.

\begin{figure}[!t]
    \includegraphics[width=\columnwidth]{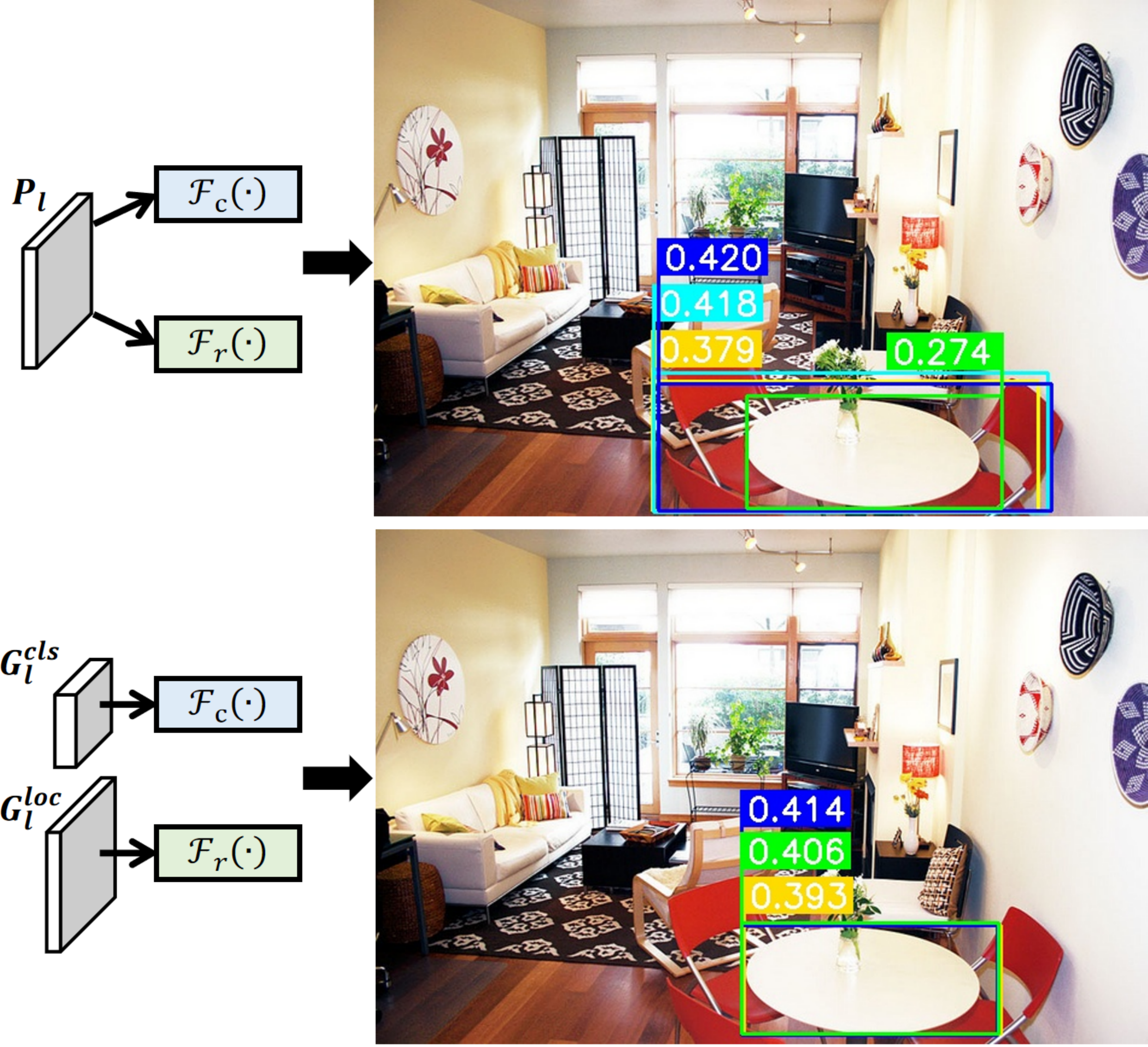}
    \caption{
    \textbf{Top:} Inference from original FCOS~\cite{tian2019fcos}. \textbf{Bottom:} Inference from FCOS with our~\ours. Results are shown before NMS. The bounding box that has the highest IoU with the ground truth is shown in green, while the top-3 bounding-boxes with the highest classification scores are marked in other colors. The competition between the two tasks in FCOS can be clearly observed in the top figure, \ie, the bounding box with the best IoU~(green) has lower classification confidence, while that with the best classification score~(blue) has a poor bounding box prediction. Thanks to our~{\ours}, the competition issue is addressed in the bottom figure, where the blue bounding box with the most confident classification prediction also has a great IoU with the ground truth.  
    }
    \label{fig:pic}
    \vspace{-5mm}
  \end{figure}

Semantic context encoded in the high-level semantic-rich features plays a crucial role in both the classification and localization task of object detection~\cite{Mottaghi_2014_CVPR,bell2016inside}.
Feature pyramid network (FPN)~\cite{lin2017feature} distills the semantic context from the high-level semantic-rich features and fuses it into the low-level detail-preserving feature maps, where small-scale objects can be better detected. This design effectively provides more semantic context for early-stage features and facilitates detecting objects in various scales.
Early works~\cite{ren2015faster,redmon2016you,he2017mask} usually tackle the detection problem by attaching a \emph{head} network shared by the two tasks on each level of feature map for a specific scale of objects, though the two tasks are semantically misaligned which is noted by Double-Head R-CNN~\cite{wu2020rethinking} and TSD~\cite{song2020revisiting} afterwards.
Typically, bounding box regression expects more texture details and edge information to accurately localize the object boundaries, while more semantic context is required to facilitate the classification of objects~\cite{dai2016r,chen2021disentangle}.
Based on this insight, they propose to use two decoupled head branches for different tasks on each feature level from FPN.
In this paper, we name this decoupling strategy the \emph{parameter decoupling}, which relies on separate heads to encode task-specific semantic context from the same feature map.

However, we observe that disentangling classification and localization only in the parameter level leads to an imperfect trade-off between two tasks.
\cref{fig:pic}~(top) visualizes the top-$3$~(for better view) bounding boxes with highest confidence scores and the box that has the highest intersection-over-union~(IoU) with corresponding ground-truth table box before non-maximum suppression~(NMS)~\cite{rothe2014non} predicted by FCOS~\cite{tian2019fcos}. FCOS has already extracted accurate enough boxes (see the green one), but they are suppressed during NMS due to relatively low confidence scores. We owe this to the intrinsic competition between the two tasks, and merely relying on the learning ability of head networks  to provide the task-specific context from a shared feature map usually shows bias towards one task, while suppressing the other.  This phenomenon has been noticed in~\cite{chen2021disentangle,song2020revisiting}, but still without proper solution being proposed. Some works~\cite{liu2018path,tan2021giraffedet,ghiasi2019fpn,tan2020efficientdet} attempt to incorporate more semantic context to improve the detection results. However, more is not always better, and there are still three problems: 1) The generality of these methods is limited and only a small number of detectors can benefit from them; 2) More computation overhead is introduced, which requires long training schedules and harms the inference efficiency; 3) More importantly, the essential problem still exists, \ie, the shared feature map is still jointly optimized for two tasks that compete with each other.

To address these issues, we propose to directly disentangle the feature encoding for classification and localization, namely \emph{\textbf{T}ask-\textbf{S}pecific \textbf{CO}ntext \textbf{DE}coupling} (\ours{} for short), so that more suitable semantic context is selected for resolving specific tasks. For the classification branch, it usually demands features with rich semantic context contained in the picture to infer the category of objects, thus we adopt spatially-coarse but semantically-strong feature encoding for it. For the localization branch, it usually requires more boundary information around objects, thus we provide it with high-resolution feature map containing more edge information for finer regression of object boundaries. Moreover, the feature fusion in two branches is designed in an efficient fashion and further boosts the performance. Benefiting from our disentangled design, incoherent context preference of the two tasks is alleviated, which allows the detector to converge faster and achieve better performance.
As shown in \cref{fig:pic}~(bottom), with \ours{}, the competition issue is addressed and the bounding box with the most confident classification prediction also has a great IoU with the ground truth.

Our main contributions are summarized as follows:
\begin{itemize}
\vspace{-8pt}\item We delve into the different roles of semantic contexts in classification and localization tasks and discover the source of their inherent conflict.
\vspace{-8pt}\item We propose a novel~\ours{} head to deal with the tangled context conflict, where different feature encodings with task-specific context can be generated. 
\vspace{-8pt}\item Extensive experiments show that \ours{} is plug-and-play and can easily boosts the performance of existing detectors with even less computational cost.

\end{itemize}

\begin{figure*}[t]
\begin{center}
    \includegraphics[width=\linewidth]{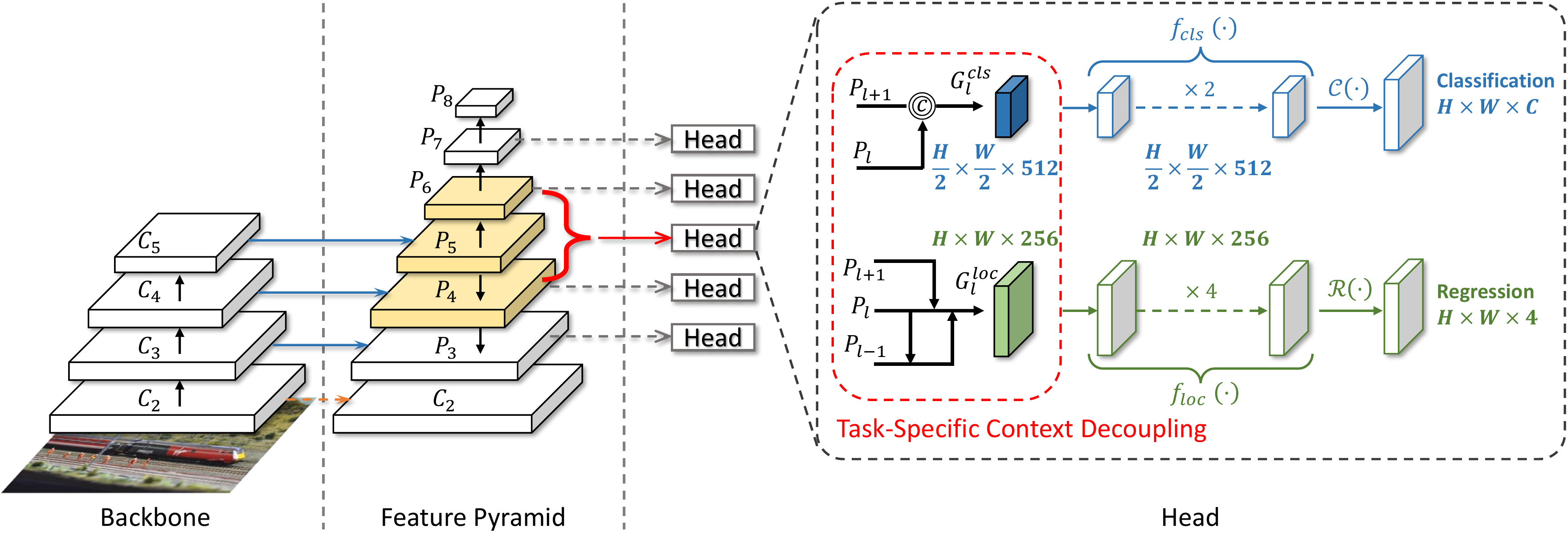}
\caption{An illustration of ours novel Task-Specific Context Decoupling~(\ours). Detector head at the $l^{th}$ pyramid level receive feature maps $P_{l+1}$,$P_{l}$ and $P_{l-1}$ from FPN~\cite{lin2017feature},~{\ours} then further disentangles the feature encoding for classification and localization tasks.}
\label{fig:pipeline}
\end{center}
\vspace{-0.6cm}
\end{figure*}


\section{Related Work}

\ptitle{Object Detection.}
The current mainstream CNN-based object detectors can be divided into two-stage~\cite{ren2015faster,he2017mask} and one-stage~\cite{redmon2016you,liu2016ssd}. Most of them use the feature pyramid network~(FPN)~\cite{lin2017feature} to cope with the large-scale variation~\cite{tan2021giraffedet,singh2018analysis} of objects. Compared with works~\cite{carion2020end,chen2021you} that merely leverage a single feature map for detecting all the scale-varying objects, FPN takes advantage of the Convolutional Neural Networks~(CNN) hierarchy, i.e., it not only fuses detail-preserving low-level features and semantic-rich high-level features, but also distributes the classification and localization tasks of different objects to corresponding feature maps according to their scale on images. While the insightful point\cite{liu2018path} that the feature maps in low layers strongly respond to edges or instance parts manifests the necessity of augmenting a bottom-up path to propagate features with strong boundary information and enhance all features with reasonable localization capability. Recent works\cite{tan2021giraffedet,tan2020efficientdet} try to stack this feature fusion structure several times for sufficient information exchange between high-level features and low-level features. These designs maybe a compromise to the inaccurate localization information contained in high-level feature maps and the insufficient semantic context contained in low-level feature maps.

\ptitle{Decoupled head.}
Decoupled head has long been the standard configuration of the one-stage detectors~\cite{lin2017focal,tian2019fcos,zhang2020bridging,li2020generalized}.
Recent works, Double-Head R-CNN~\cite{wu2020rethinking} and TSD~\cite{song2020revisiting} revisit the specialized sibling head that is widely used in R-CNN family~\cite{girshick2014rich,girshick2015fast,ren2015faster,cai2018cascade,he2017mask,pang2019libra} and finally figure out the essence of the tasks misalignment between classification and localization. YOLOX~\cite{ge2021yolox} also points out that the coupled detection head may harm the performance, it introduces decoupled head to the YOLO family~\cite{redmon2016you,redmon2017yolo9000,redmon2018yolov3} for the first time and greatly improves the converging speed and boosts the permanformance. Base on the decoupled head, DDOD~\cite{chen2021disentangle} proposes to use deformable convolutions to learn separate convolutional offset for each branch, aiming to adaptively select specific spatial features for each head. These works demonstrate the importance of decoupling between classification and localization tasks. However, as mentioned in section~\ref{sec:intro}, the decoupling for classification and localization only in the parameter level leads to an imperfect trade-off between two tasks.


\section{Method}

\subsection{Motivation and Framework}
\label{sec:motivation}

Classification and localization are two highly related but still contradictory tasks in object detection.
For each object, classification is more coarse-grained which requires richer semantic context, while localization is rather fine-grained and demands more on detailed boundary information.
For this reason, mainstream detectors~\cite{lin2017focal, zhang2020bridging, li2020generalized, ge2021ota, song2020revisiting, wu2020rethinking} apply decoupled head to cope with this conflict. Specifically, given an ground-truth object assigned to a specific pyramid level $l$, with bounding box $\mathcal{B}$ and class label ${c}$, the detectors with the typical decoupled head minimize the classification and localization loss based on the same feature map $P_l$:
\begin{equation}
\begin{split}
\mathcal{L} = \mathcal{L}_{cls}(\mathcal{F}_c (P_l), c) + \mathcal{L}_{loc}(\mathcal{F}_r (P_l), \mathcal{B}),
\label{L1}
\end{split}
\end{equation}
where $\mathcal{F}_c(\cdot) \tight{=} \{f_{cls}(\cdot), \mathcal{C}(\cdot)\}$, $\mathcal{F}_r(\cdot) \tight{=} \{f_{loc}(\cdot), \mathcal{R}(\cdot)\}$, are the classification and localization branches.
$f_{cls}(\cdot)$ and $f_{loc}(\cdot)$ are the feature projection functions for classification and localization, while $\mathcal{C}(\cdot)$ and $\mathcal{R}(\cdot)$ are the final layers in two branches which decode features to classification scores and bounding box positions.
In the common decoupled-head design, $f_{cls}(\cdot)$ and $f_{loc}(\cdot)$ share the same structure but are learned with seperate parameters to provide each task with different feature contexts, \ie, \emph{parameter decoupling}.
However, this simplistic design cannot fully solve this problem as the semantic context has been largely determined by the shared input features $P_l$.
Although recent work~\cite{chen2021disentangle} attempts to learn features with more flexible context for each task with deformable convolutions, the fact that they still originate from the same features, however, limits its effectiveness.
Therefore, the conflict between classification and localization imposes opposite preferences of context in $P_l$, leading to an imperfect balance between the two tasks.

To address this issue, our TSCODE decouples the feature encoding for the two tasks at the source and leverages feature maps with different semantic context in the two branches.
Instead of using $P_l$ as the common input, we feed the two branches with task-specific input features, \ie, $G^{cls}_l$ and $G^{loc}_l$. To this end, \cref{L1} can be written as:
\begin{equation}
\begin{split}
\mathcal{L} = \mathcal{L}_{cls}(\mathcal{F}_c (G^{cls}_l), c) + \lambda\mathcal{L}_{loc}(\mathcal{F}_r (G^{loc}_l), \mathcal{B}).
\label{L2}
\end{split}
\end{equation}
For the classification branch, we generate spatially coarser but semantically richer feature maps.
While for the localization branch, we provide it with feature maps contain more detailed texture and boundary information.

As illustrated in~\cref{fig:pipeline}, our method follows the common one-stage detection framework, which is composed of the backbone, the feature pyramid and the detection head.
The backbone and feature pyramid generate multi-scale feature maps from the input images.
Our~\ours{} head then receives three levels of feature maps and generate decoupled feature maps for classification and localization.
More importantly, \ours{} is \emph{plug-and-play} and can be easily incorperated into most of the popular detectors, whether it is anchor-based or anchor-free.

\subsection{Semantic Context Encoding for Classification}

In object detection, classification is a relatively coarse-grained task which recognizes \emph{what} an bounding box covers. On the one hand, as mentioned in~\cite{jiang2018acquisition,feng2021tood,song2020revisiting,chen2021disentangle}, the classification can often be pinned down by the features of its key or salient parts of an object, \ie, \emph{local focus}.
However, existing methods ignore that the salient areas could be sparsely distributed, indicating that there could be feature redundancy in the classfication branch.
We find in the experiments that, feeding the classification branch with downsampled feature maps witnesses almost no performance drop, but reduces the computational cost (results can be seen in \cref{sec:exp-ablation}). On the other hand, the category of an object could be infered from its surrounding environment, \eg, a chair is more likely to co-occur with a table, or an occluded table can be recognized from a larger region. This means the \emph{global abstraction} from a richer semantic context could facilitate the classification task.

\begin{figure}[t]
\includegraphics[width=\columnwidth]{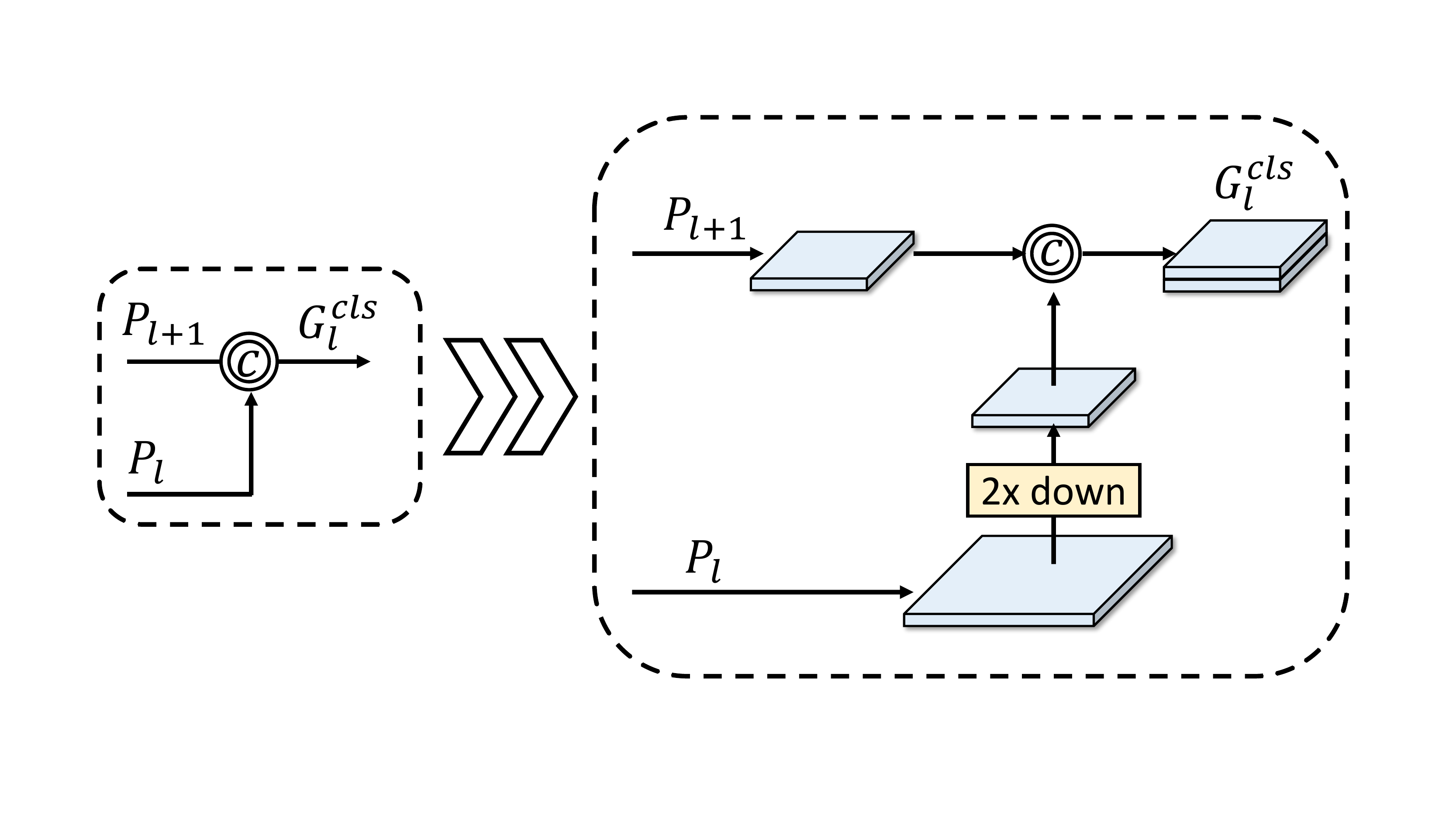}
\caption{
Semantic context encoding for classification.
}
\label{fig:SCE}
\vspace{-0.2cm}
\end{figure}

Based on these insights, we devise \emph{Semantic Context Encoding} (SCE) for efficient and accurate classification. Specifically, at each pyramid level $l$, SCE leverages the feature maps from two levels, \ie, $P_{l}$ and $P_{l+1}$, to generate a semantically-rich feature map for classification.
As shown in~\cref{fig:SCE}, we first downsample $P_l$ by a factor of $2$ and concatenate it with $P_{l+1}$ to generate the final $G^{cls}_l$:
\begin{equation}
G^{cls}_l = \mathtt{Concat}\big(\mathtt{DConv}(P_l),P_{l+1}\big),
\end{equation}
where $\mathtt{Concat}(\cdot)$ and $\mathtt{DConv}(\cdot)$ represent concatenation and a shared downsampling convolutional layer.
Note that $G^{cls}_l$ is in the $1/2$ resolution of $P_{l}$.
$G^{cls}_l$ is then passed into $\mathcal{F}_c(\cdot)$ = $\{f_{cls}(\cdot), \mathcal{C}(\cdot)\}$ to predict the classification scores.
Instead of using $4$ convolutional layers with $256$ channels, we adopt a shallow-but-wide design of $f_{cls}(\cdot)$ using $2$ convolutional layers with $512$ channels.
We argue that this design can encode more semantic information than the deep-but-narrow one, which requires little extra computational cost but facilitates more accurate classification.
Since $G^{cls}_l$ is $2{\times}$ downsampled than $P_l$, each location $(x, y)$ in $G^{cls}_l$ predicts the classification scores of its four nearest neighbors in $P_l$, denoted as $\tilde{C} \in \mathbb{R}^{H_{l+1} \times W_{l+1} \times 4N}$, and $N$ is the number of categories. $\tilde{C}$ is then rearranged to $\hat{C} \in \mathbb{R}^{H_{l} \times W_{l} \times N}$ to recover the resolution:
\begin{align}
\hat{C}[2x\!+\!i,2y\!+\!j,c]\!=\!\tilde{C}[x,y,(2i\!+\!j)c],\forall i,j\!\in\!\{0,\!1\}.
\end{align}
By this way, we can not only leverage the sparsity of the salient features (from $P_{l}$), but also benefit from the rich semantic context in higher pyramid level (from $P_{l+1}$). This helps infer object categories more effectively, especially for those texture-less objects or those with severe occlusion.

\begin{figure}[t]
\includegraphics[width=\columnwidth]{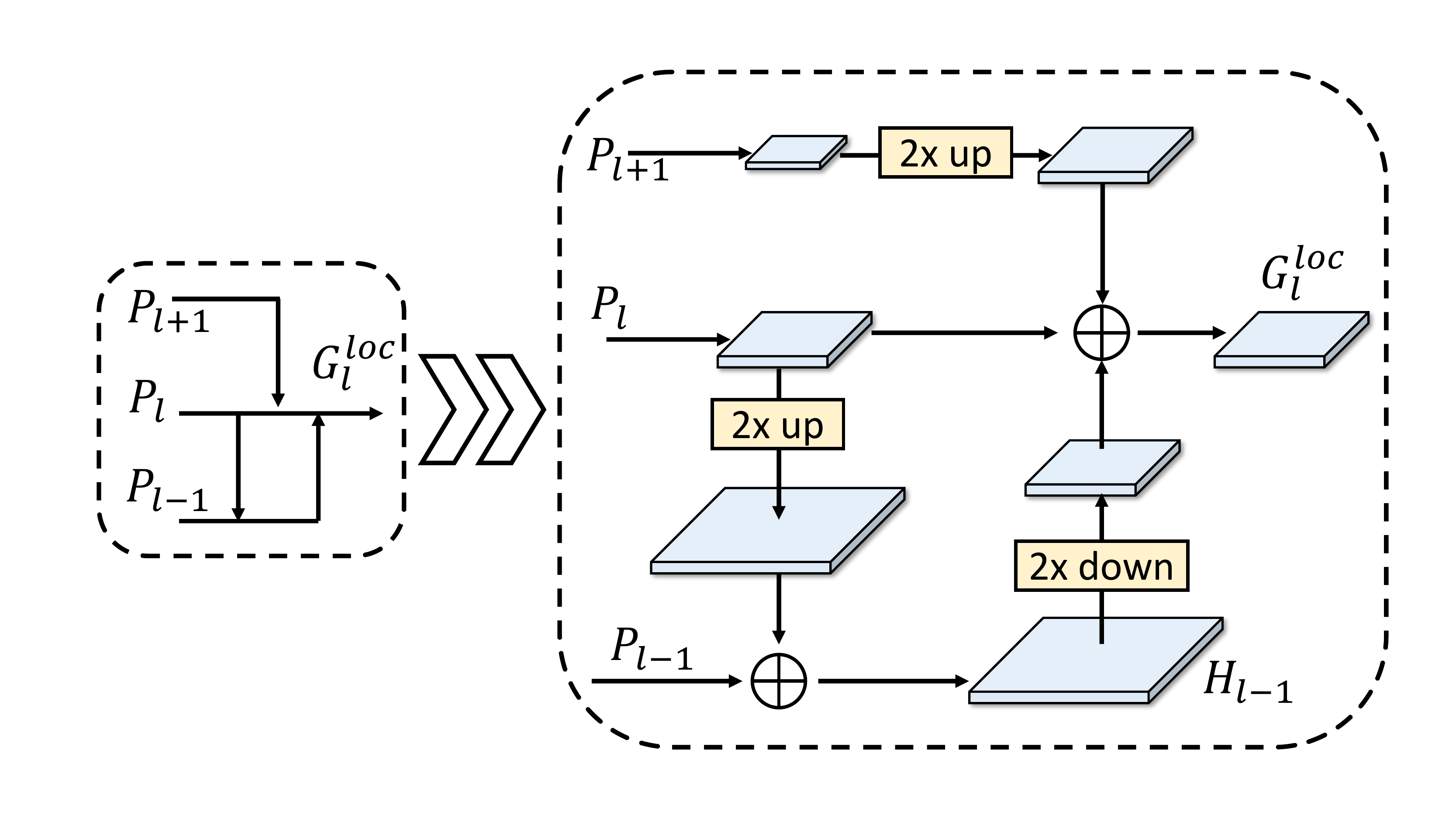}
\caption{Detail-preserving encoding for localization.}
\label{fig:DPE}
\vspace{-0.2cm}
\end{figure}

\subsection{Detail-Preserving Encoding for Localization}

Unlike classification, localization is a more fine-grained task which relies on more texture details and boundary information to predict the corners of an object. However, existing methods usually regress the object corners from a single-scale feature map $P_l$. The feature maps in lower pyramid levels have stronger response to contour, edge and detailed texture of objects. This can further benefit the localization task but often comes with huge extra computational cost~\cite{liu2018path,yang2022querydet}. QueryDet~\cite{yang2022querydet} uses sparse convolution~\cite{graham20183d} to reduce computational cost on low-level feature maps, but it still requires an extra auxiliary branch with specific supervision signal.
On the other hand, we further note that high-level feature maps are also important for localization as it helps see the whole object as completely as possible, which provides more information to infer the overall shape of objects.

Based on these observations, we propose \emph{Detail-Preseving Encoding} (DPE) to achieve accurate localization without sacrificing efficiency. In each pyramid level $l$, our DPE accepts the feature maps from three pyramid levels, \ie, $P_{l-1}$, $P_{l}$ and $P_{l+1}$. $P_{l-1}$ provides more detail and edge features while $P_{l+1}$ provides a more thorough perspective of objects. We demonstrate that each pyramid level is mainly related to the two neighboring levels and further levels could even harm the performance.
The structure of DPE is shown in~\cref{fig:DPE}. 
For computational efficiency, we adopt a simplistic U-Net~\cite{ronneberger2015u} to fuse $P_{l-1}$ and $P_{l+1}$.
$P_{l}$ is first upsampled by a factor of $2$ and then aggragated with $P_{l-1}$. And a $3 \times 3$ convolutional layer with a stride of $2$ downsamples it to the resolution of $P_l$. This design effectively preserves the detail information in $P_{l-1}$ with little extra computational cost.
At last, $P_{l+1}$ is upsampled and aggregated to generate the final $G^{loc}_{l}$.
The computation can be written as:
\begin{align}
\begin{split}
G^{loc}_{l} = P_l + \mu(P_{l+1}) + \mathtt{DConv}(\mu(P_l) + P_{l-1})
\end{split}&
\end{align}
where $\mu(\cdot)$ represents upsampling and $\mathtt{DConv}(\cdot)$ is another shared downsampling convolutional layer.
Speically, we compute $G^{loc}_3$ with $C_2$, $P_3$ and $P_4$, as computing $P_2$ through FPN induces huge computational cost.
Afterwards, $\mathcal{F}_r(\cdot) = \{f_{los}(\cdot), \mathcal{R}(\cdot)\}$ further predicts the bounding boxes in the $l^{\text{th}}$ pyramid level based on $G^{loc}_l$.

\begin{table*}[t]
\small
  \begin{center}
    \begin{tabular}{l|cc|lcc|ccc|ccc|c}
      \toprule[1pt]
      Method & SCE      & DPE     &         AP        &        AP$_{50}$      &AP$_{75}$& AP$_{S}$&AP$_{M}$&AP$_{L}$&AR$_{S}$&AR$_{M}$&AR$_{L}$& GFLOPs   \\
      \hline
      \hline
\multirow{4}*{FCOS\cite{tian2019fcos}}          &      &      & 38.7     & 57.4      & 41.8    & 22.9    & 42.5   & 50.1   & 36.8   & 61.7   & 73.2   & 200.59      \\
             & \checkmark &            &  39.3$_{\textcolor{green}{+0.6}}$  & 58.2      & 42.8    & 23.3    & 43.3   & 50.8   & 37.5   & 62.1   & 73.2   & 182.62      \\
             &            & \checkmark &  38.9$_{\textcolor{green}{+0.2}}$  & 57.5      & 41.8    & 22.9    & 42.8   & 50.3   & 37.0   & 62.4   & 73.2   & 213.19      \\
             & \checkmark & \checkmark &  {\textbf{40.0}}$_{\textcolor{green}{+1.3}}$  & {\textbf{58.7}}      & {\textbf{43.1}}    & 23.7    & 44.0   & 51.8   & 38.0   & 62.7   & 73.3   & 195.22      \\
      \hline
\multirow{4}*{ATSS\cite{zhang2020bridging}}     &      &      & 39.4     & 57.6      & 42.8    & 23.6    & 42.9   & 50.3   & 38.2   & 63.5   & 73.6   & 205.30      \\
             & \checkmark &            &  40.2$_{\textcolor{green}{+0.8}}$  & 58.6      & 43.8    & 23.9    & 44.0   & 52.2   & 38.9   & 63.7   & 74.0   & 187.35      \\
             &            & \checkmark &  39.8$_{\textcolor{green}{+0.4}}$  & 57.8      & 42.9    & 23.7    & 43.1   & 50.9   & 38.0   & 64.0   & 75.3   & 217.89      \\
             & \checkmark & \checkmark &  {\textbf{40.8}}$_{\textcolor{green}{+1.4}}$  & {\textbf{59.0}}      & {\textbf{44.4}}    & 23.7    & 44.6   & 52.7   & 38.7   & 64.4   & 75.4   & 199.94      \\
      \hline
\multirow{4}*{GFL\cite{li2020generalized}}     &      &      & 40.2     & 58.4      & 43.3    & 23.3    & 44.0   & 52.2   & 38.0   & 62.9   & 74.1   & 208.39      \\
             & \checkmark &            &  41.1$_{\textcolor{green}{+0.9}}$  & 59.3      & 44.5    & 24.0    & 44.6   & 53.8   & 37.7   & 63.4   & 75.2   & 190.44      \\
             &            & \checkmark &  40.6$_{\textcolor{green}{+0.4}}$  & 58.4      & 43.8    & 23.5    & 44.2   & 53.2   & 37.2   & 63.9   & 74.6   & 220.99      \\
             & \checkmark & \checkmark &  {\textbf{41.6}}$_{\textcolor{green}{+1.4}}$  & {\textbf{59.8}}      & {\textbf{44.9}}    & 23.8    & 45.3   & 54.8   & 38.4   & 64.2   & 75.0   & 203.04      \\
     \bottomrule[1pt]
    \end{tabular}
    \end{center}
\vspace{-3mm}
\caption{Ablation studies on COCO \textit{mini-val} set. {\textbf {SCE}} stands for using of Semantic Context Encoding on classification; {\textbf {DPE}} stands for using of Detail-Preserving Encoding on localization.}

\label{coco-ablation}
\vspace{-1mm}
\end{table*}
\section{Experiments}

\ptitle{Dataset and Evaluation Metric.}
Our experiments are conducted on the large-scale detection benchmark MSCOCO 2017~\cite{lin2014microsoft}.
Following common practice~\cite{tian2019fcos,zhang2020bridging,li2020generalized},
we use COCO \textit{trainval35k} split~($115$K images) for training and \textit{minival} split~($5$K images) for validation.
We report our main results and compare with previous detectors on the \textit{test-dev} split~($20$K images)
by uploading the detection results to the evaluation server.

\ptitle{Implementation and Training Details.}
We implement our~\ours{} with MMDetection~\cite{chen2019mmdetection} and cvpods~\cite{zhu2020cvpods}.
Unless otherwise noted, we utilize a ResNet-50~\cite{he2016deep} backbone network which is pretrained on ImageNet~\cite{deng2009imagenet}. As~\ours{} can be applied as a plug-and-play head to other basic detectors, we follow the original settings in the basic detectors~(including loss function, bounding box parameterization, label assignment strategy and hyper-parameters) and merely replace the head part with \ours{}.
The input images are resized to a maximum scale of $1333 \times 800$ during testing without changing the aspect ratio.
The computational cost (measured in FLOPs) is calculated under the input size of $1280 \times 800$.
We set the batch size to $16$ and use $8$ V100 GPUs~($2$ images per GPU) to train our models. Following common practice, we apply the standard $1{\times}$ training schedule in the ablation studies. The learning rate starts from $0.01$ and decayed by a factor of $10$ after $8$ and $11$ epochs, respectively. We also compare our method with state-of-the-art detectors with different backbone networks, where we adopt $2{\times}$ training schedule and multi-scale training. Please refer to~\cref{cmp_sota} for more details.

\subsection{Ablation Studies}
\label{sec:exp-ablation}

We first conduct extensive ablation studies to evaluate the efficacy of our design on COCO \emph{minival}. We use ResNet-$50$ as the backbone network and all the models are trained for $12$ epochs following $1{\times}$ schedule.

\begin{figure}[!t]
    \includegraphics[width=\columnwidth]{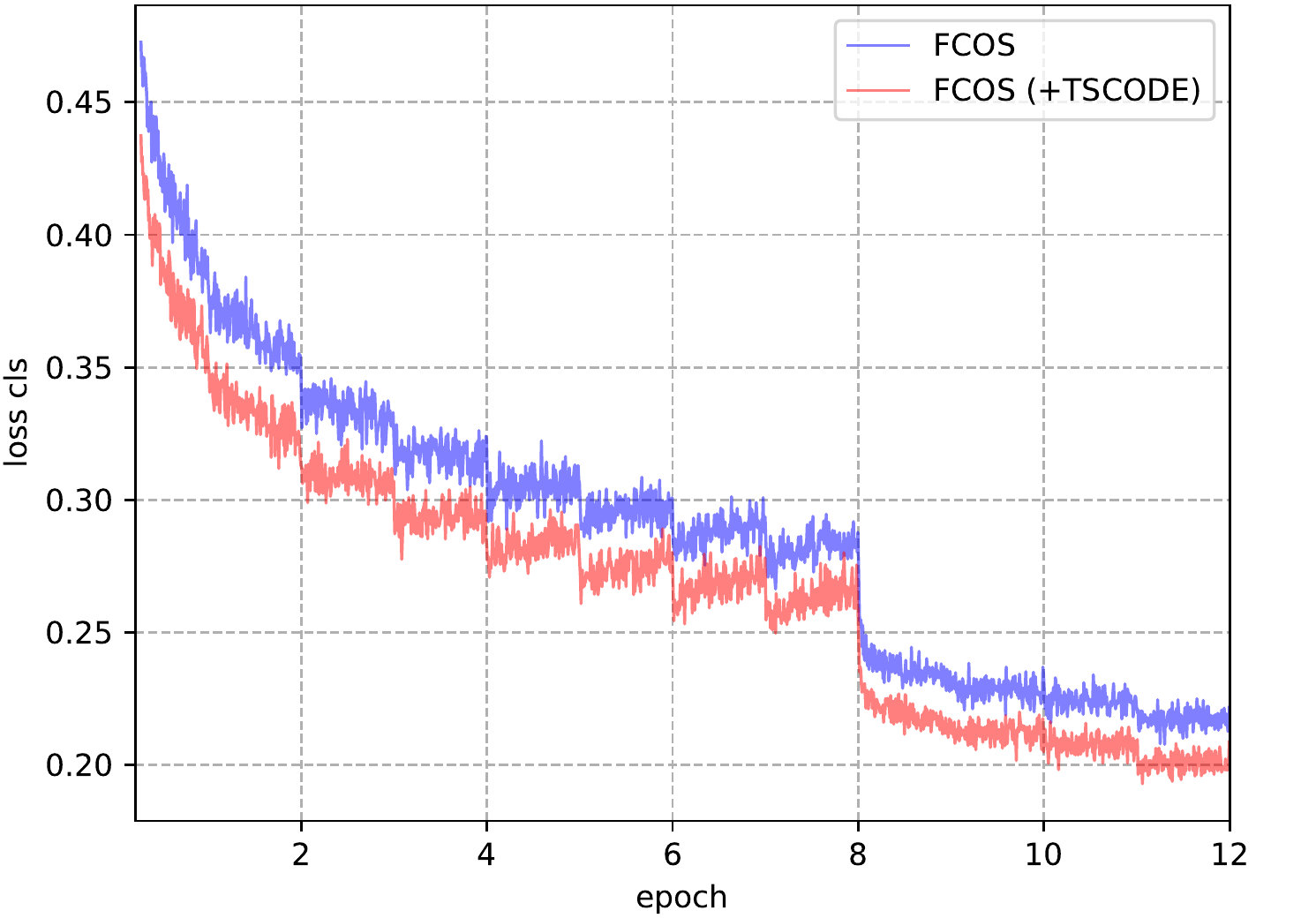}
    \caption{
    Comparison of classification training loss between FCOS with and without \ours{}. \ours{} can accelerate the training and contributes to better convergence.
    }
    \label{loss}
    \vspace{-0.2cm}
  \end{figure}

\ptitle{Effects of Individual Components.}
We first study the effectiveness of each component of \ours{}. In the experiments, we ablate our method with three basic detectors, \ie, FCOS~\cite{tian2019fcos}, ATSS~\cite{zhang2020bridging} and GFL~\cite{li2020generalized}, to evaluate the generality of our designs. For fair comparison, we retrain the baseline models strictly following the original papers with similar or slightly better results.

As demonstrated in~\cref{coco-ablation}, applying only Semantic Context Encoding~(SCE) improves the AP of the baseline models by $0.6{\sim}0.9$ points, while reducing the computational cost (FLOPs) by $9\%$.
And note that more significant improvements are observed for large and medium objects.
This is consistent with our motivation to leverage more semantic context for the classification task.
Moreover, better classification accuracy also facilitates find more objects, and thus SCE achieves better AR results on all the basic detectors.

When using Detail-Preserving Encoding~(DPE) alone, we also observe consistent improvements on all basic detectors, especially on large objects. However, the gains are less than those from SCE, indicating that the feature context in the original decoupled-head design tends to lean to the localization task.

At last, we observe significant improvements on all metrics with our full model. The models with~\ours{} consistently outperforms the baseline models by $1.3{\sim}1.4$ AP points, which is higher than the sum of the individual improvements. Benefitting from the decoupled feature contexts, the classification branch can learn richer semantic information to infer the category of objects, while the localization branch can benefit from more edge details to accurately predict the object boundaries. Moreover, our method also slightly reduces the computational cost, demonstrating the good efficiency of our design.
We further visualize the classification loss when training FCOS~\cite{tian2019fcos} with and without \ours{} in \cref{loss}. \ours{} can accelerate the training and contributes to better convergence. And similar results can be observed in other basic detectors.

\begin{table}[tb]

\small
\centering
\begin{tabular}{c|c|c|ccc|c}
\toprule[1pt]
Layer & Kernel & $P_{l+1}$ & AP & AP$_{50}$ & AP$_{75}$ & GFLOPs \\
\hline
\hline
- & - &  & 38.7 & 57.4 & 41.8 & 200.59 \\
\hline
conv & 3 $\times$ 3 &  & 38.6 &  57.3 &  41.8 & 165.99 \\
\hline
avg & \multirow{3}{*}{3 $\times$ 3} & \checkmark & 39.1 & 57.8 & 42.7 & 179.47 \\
max &  & \checkmark & 39.1 & 57.9 & 42.5 & 179.47 \\
conv &  & \checkmark & \textbf{39.3} & 58.2 & 42.8 & 182.62 \\
\hline
\multirow{3}{*}{conv} & 3 $\times$ 3 & \checkmark & \textbf{39.3} & 58.2 & 42.8 & 182.62 \\
 & 5 $\times$ 5 & \checkmark & 39.2 & 58.1 & 42.8 & 188.23 \\
 & 7 $\times$ 7 & \checkmark & 39.2 & 58.2 & 42.3 & 196.65 \\
\bottomrule[1pt]
\end{tabular}

\vspace{-1mm}
\caption{Performance of different ways to generate Semantic Context Encoding~(SCE) for classification branch on FCOS~\cite{tian2019fcos}.}
\label{SCE}
\vspace{-3mm}
\end{table}

\ptitle{Semantic Context Encoding.}
Next, we further study the influence of different ways to conduct SCE with FCOS.
The results are shown in \cref{SCE}.
We first feed the classification branch with merely the downsampled $P_{l}$. It is observed that this model achieves similar results with the baseline but with significantly less computation, indicating the feature redundancy in the classification branch.
Next, we put the high-level $P_{l+1}$ into SCE and vary the operation to downsample $P_{l}$. The kernel size is fixed to $3 \times 3$ and the stride is $2$ for all operations. As observed in \cref{SCE}, average pooling and max pooling performs slightly worse than convolution, albeit with less computational cost. We assume that average pooling is inefficient in extracting sharp classification signals, while the sparse connections caused by max pooling harms the convergence of the models.
At last, we study the influence of the kernel size in convolutions. And a large kernel does not necessarily brings better results. We argue that a large kernel may include too much noise signals, which harms the classification performance. However, the performance of the detector is still improved regardless of how SCE is generated.

\begin{table}[tb]
\centering
\setlength{\tabcolsep}{1.mm}
\resizebox{\columnwidth}{!}{
\begin{tabular}{cccc|ccc|ccc}
\toprule[1pt]
$P_{l+2}$  &$P_{l+1}$  & $P_l$      & $P_{l-1}$  & AP   & AP$_{50}$ & AP$_{75}$ & AP$_S$   & AP$_M$ & AP$_L$ \\
\hline\hline
		   &		   & \checkmark &            & 39.3 & 58.2      & 42.8      & 23.3 & 43.3 	   & 50.8      \\
\hline
		   &\checkmark & \checkmark &            & 39.4 & 58.3      & 42.6      & 23.3 & 43.2      & 51.0      \\
		   &\checkmark &            & \checkmark & 39.6 & 58.4      & 42.8      & 23.9 & 43.3      & 51.4      \\
		   &           & \checkmark & \checkmark & 39.7 & 58.1      & 43.1      & 23.2 & 43.7      & 51.8      \\
		   &\checkmark & \checkmark & \checkmark & \textbf{40.0} & 58.7      & 43.1      & 23.7 & 44.0      & 51.8      \\
\checkmark &\checkmark & \checkmark & \checkmark & 39.9 & 58.6      & 43.2      & 23.1 & 43.7      & 52.1      \\
\bottomrule[1pt]
\end{tabular}
}
\vspace{-1mm}
\caption{Performance of different ways to generate Detail-Preserving Encoding~(DPE) for localization branch on FCOS~\cite{tian2019fcos}.}
\label{DPE}
\end{table}

\ptitle{Detail-Preserving Encoding.}
At last, we investigate the efficacy of our design of DPE by ablating the feature maps from different levels in~\cref{DPE}, where SCE is used in all the models by default.
Individualy aggregating the high-level feature map $P_{l+1}$ with $P_{l}$ only achieves marginal improvements, but the incorperation of $P_{l-1}$ contributes to more significant performance gains as more detail and edge information is obtained. Interestingly, the model with $P_{l+1}$ and $P_{l-1}$ outperforms the one with $P_{l+1}$ and $P_{l}$, which again demonstrates the importance of detail information. And the model with all three feature maps achieves significant improvements, especially on large and medium objects. Note that the gains from three levels are greater than the sum of the individual gains, which means $P_{l+1}$ could provide more benefit if there are adequate detail information. At last, we further add $P_{l+2}$ and find a slight decay in the results, suggesting that too much environmental information may be useless or even harmful for bounding box regression.

\begin{table}[tb]
\centering
\setlength{\tabcolsep}{1.mm}
\resizebox{\columnwidth}{!}{
\begin{tabular}{l|lcc|ccc}
\toprule[1pt]
Method & AP & AP$_{50}$ & AP$_{75}$ & AP$_{S}$ & AP$_{M}$ & AP$_{L}$ \\
\hline\hline
FCOS~\cite{tian2019fcos} & 38.7 & 57.4 & 41.8 & 22.9 & 42.5 & 50.1 \\
FCOS (+ours) & 40.0$_{\textcolor{green}{+1.3}}$ & 58.7 & 43.1 & 23.7 & 44.0 & 51.8 \\
\hline
ATSS~\cite{zhang2020bridging} & 39.4 & 57.6 & 42.8 & 23.6 & 42.9 & 50.3 \\
ATSS (+ours) & 40.8$_{\textcolor{green}{+1.4}}$ & 59.0 & 44.4 & 23.7 & 44.6 & 52.7 \\
\hline
GFL~\cite{li2020generalized} & 40.2 & 58.4 & 43.3 & 23.3 & 44.0 & 52.2 \\
GFL (+ours) & 41.6$_{\textcolor{green}{+1.4}}$ & 59.8 & 44.9 & 23.8 & 45.3 & 54.8 \\
\hline
AutoAssign~\cite{zhu2020autoassign} & 40.4 & 59.6 & 43.7 & 22.7 & 44.1 & 52.9 \\
AutoAssign (+ours) & 41.1$_{\textcolor{green}{+0.7}}$ & 60.2 & 44.1 & 23.0 & 45.0 & 54.2 \\
\hline


DDOD~\cite{chen2021disentangle} & 41.6 & 59.9 & 45.2 & 23.9 & 44.9 & 54.4 \\
DDOD (+ours) & 42.4$_{\textcolor{green}{+0.8}}$ & 60.2 & 46.3 & 24.5 & 45.5 & 56.0 \\
\hline
DeFCN$^*$~\cite{wang2021end} & 41.4 & 59.5 & 45.6 & 26.1 & 44.9 & 52.0 \\
DeFCN$^*$ (+ours) & 42.3$_{\textcolor{green}{+0.9}}$ & 60.7 & 46.7 & 27.0 & 45.5 & 53.6\\
\hline
OTA~\cite{ge2021ota} & 40.7 & 58.4 & 44.3 & 23.2 & 45.0 & 53.7 \\
OTA (+ours) & 41.5$_{\textcolor{green}{+0.8}}$ & 59.2 & 45.3 & 23.4 & 45.2 & 55.7 \\
\hline

DW\cite{li2022dual} & 41.5 & 59.8 & 44.8 & 23.4 & 44.9 & 54.8 \\
DW (+ours) & 42.0$_{\textcolor{green}{+0.5}}$ & 60.3 & 45.1 & 24.1 & 45.5 & 56.0 \\
\bottomrule[1pt]
\end{tabular}
}
\vspace{-2mm}
\caption{Applying~{\ours} into various popular dense object detectors. Method with $^*$ means training for 3$\times$ schedule follow its official repository.}
\vspace{-3mm}
\label{more}
\end{table}

\begin{table*}[t]
\small
    \renewcommand\arraystretch{1.2}
    \centering
    \vspace{-2pt}
        {
            \begin{tabular}{lcccccccc}
                \toprule
                Method  & Backbone & AP &AP$_{50}$ &AP$_{75}$ &AP$_{S}$ &AP$_{ M}$ &AP$_{ L}$ & Reference\\
                \cmidrule(r){1-1}
                \cmidrule(r){2-2}
                \cmidrule(r){3-5}
                \cmidrule(r){6-8}
                \cmidrule(r){9-9}

                ATSS~\cite{zhang2020bridging}& ResNet-101  &43.6 &62.1 &47.4 &26.1 &47.0 &53.6 & CVPR20 \\
                PAA~\cite{kim2020probabilistic} & ResNet-101  & 44.8 & 63.3 & 48.7 & 26.5 & 48.8 & 56.3 & ECCV20\\
                GFL \cite{li2020generalized}&ResNet-101 & 45.0 & 63.7 & 48.9 & 27.2 & 48.8 & 54.5 & NeurIPS20 \\
                GFLV2\cite{li2021generalized} &ResNet-101  & 46.2 & 64.3 & 50.5 & 27.8 & 49.9 & 57.0 & CVPR21 \\
                OTA~\cite{ge2021ota} & ResNet-101  & 45.3 & 63.5 & 49.3 & 26.9 & 48.8 & 56.1 & CVPR21\\
                IQDet~\cite{ma2021iqdet} & ResNet-101  & 45.1 & 63.4 & 49.3 & 26.7 & 48.5 & 56.6 & CVPR21\\
                ObjectBox~\cite{zand2022objectbox} & ResNet-101 & 46.1 & 65.0 & 48.3 & 26.0 & 48.7 & 57.3 & ECCV22\\
                \ours{}~\textbf{(ours)} +~\cite{li2020generalized} &ResNet-101  & \textbf{46.7} & \textbf{65.1} & \textbf{50.8} & \textbf{28.2} & \textbf{50.3} & \textbf{57.7} & --\\
                \cmidrule(r){1-1}
                \cmidrule(r){2-2}
                \cmidrule(r){3-5}
                \cmidrule(r){6-8}
                \cmidrule(r){9-9}

                ATSS~\cite{zhang2020bridging}& ResNeXt-101-32x8d  &45.1 &63.9 &49.1 &27.9 &48.2 &54.6 & CVPR20 \\
                PAA~\cite{kim2020probabilistic} & ResNeXt-101-64x4d  & 46.6 & 65.6 & 50.8 & 28.8 & 50.4 & 57.9 & ECCV20\\
                GFL~\cite{li2020generalized}&ResNeXt-101-32x4d  & 46.0 & 65.1 & 50.1 & 28.2 & 49.6 & 56.0 & NeurIPS20 \\
                GFLV2~\cite{li2021generalized} &ResNeXt-101-32x4d  & 47.2 & 65.7 & 51.7 & 29.1 & 50.8 & 58.2 & CVPR21 \\
                OTA~\cite{ge2021ota} & ResNeXt-101-64x4d & 47.0 & 65.8 & 51.1 & 29.2 & 50.4 & 57.9 & CVPR21\\
                IQDet~\cite{ma2021iqdet} & ResNeXt-101-64x4d  & 47.0 & 65.7 & 51.1 & 29.1 & 50.5 & 57.9 & CVPR21\\
                \ours~\textbf{(ours)} +~\cite{li2020generalized} &ResNeXt-101-32x4d & 47.6 & 66.3 & 51.8 & 29.5 & 51.1 & 58.6 & --\\
                \ours~\textbf{(ours)} +~\cite{li2020generalized} &ResNeXt-101-64×4d & \textbf{48.3} & \textbf{67.1} & \textbf{52.6} & \textbf{30.0} & \textbf{51.9} & \textbf{59.9} & --\\
                \cmidrule(r){1-1}
                \cmidrule(r){2-2}
                \cmidrule(r){3-5}
                \cmidrule(r){6-8}
                \cmidrule(r){9-9}
                
                ATSS~\cite{zhang2020bridging}& ResNeXt-101-32x8d-DCN  & 47.7 & 66.6 &52.1 & 29.3 &50.8 & 59.7 & CVPR20 \\
                PAA~\cite{kim2020probabilistic} & ResNeXt-101-64x4d-DCN  & 49.0 & 67.8 & 53.3 & 30.2 & 52.8 & 62.2 & ECCV20\\
                GFL~\cite{li2020generalized}&ResNeXt-101-32x4d-DCN  & 48.2 & 67.4 & 52.6 & 29.2 & 51.7 & 60.2& NeurIPS20 \\
                GFLV2~\cite{li2021generalized} &ResNeXt-101-32×4d-DCN  & 49.0 & 67.6 & 53.5 & 29.7 & 52.4 & 61.4 & CVPR21 \\
                OTA~\cite{ge2021ota} & ResNeXt-101-64x4d-DCN  & 49.2 & 67.6 & 53.5 & 30.0 & 52.5 & 62.3 & CVPR21\\
                IQDet~\cite{ma2021iqdet} & ResNeXt-101-64x4d-DCN  & 49.0 & 67.5 & 53.1 & 30.0 & 52.3 & 62.0 & CVPR21\\
                \ours~\textbf{(ours)} +~\cite{li2020generalized} &ResNeXt-101-32×4d-DCN & 50.0 & 68.5 & 54.6 & 31.0 & 53.4 & 62.6 & --\\
                \ours~\textbf{(ours)} +~\cite{li2020generalized} &ResNeXt-101-64×4d-DCN & \textbf{50.8} & \textbf{69.3} & \textbf{55.3} & \textbf{31.7} & \textbf{54.2} & \textbf{63.7} & --\\
                \bottomrule
            \end{tabular}
        }
    \vspace{-6pt}
    \caption{ Performance comparison with state-of-the-art detectors \emph{(single-model and single-scale results except the last row)} on COCO {\tt test-dev}. {\ours} consistently outperforms the strong baseline GFL~\cite{li2020generalized} by $1.6\sim1.8$~AP and even outperforms its improved version GFLV2~\cite{li2021generalized}. \textbf{DCN}: Deformable Convolutional Network \cite{dai2017deformable}. }
    \vspace{-6pt}
    \label{sota}
\end{table*}

\subsection{Generality to Different Detectors}
To evaluate the generality of~\ours{}, we further apply it to $8$ recent popular one-stage detectors~\cite{zhu2020autoassign,chen2021disentangle,wang2021end,ge2021ota,li2022dual,tian2019fcos,li2020generalized,zhang2020bridging} and evaluate the performance on the COCO \textit{minival}.
As shown in \cref{more}, \ours{} consistently improves the performance of different baseline detectors by $0.5{\sim}1.4$ AP points. 
Notably, \ours{} achieves improvements of $0.8$ AP on the detector DDOD~\cite{chen2021disentangle} which is designed with a specific disentanglement structure for the classification and localization tasks. Moreover, our method provides the latest detector DW~\cite{li2022dual} with a performance gain of $0.5$ AP, which further demonstrates the efficacy of~\ours{}.

Moreover, DPE can also be regarded as a simple feature fusion module if its output features are directly sent to the classification branch~(without context decoupling). To this end, we compare it with the popular PAFPN~\cite{liu2018path}. They differ in two aspects: 
First, we leverage a semantic context augmentation from $P_{l}$ to $P_{l-1}$ before bottom-up aggregation ($P_{l-1}$ back to $P_{l}$) which can enrich the feature representation and benifit the classification and localization tasks.
Second, the weight of convolution downsampling is shared between all the head, which not only makes the detector parameter-efficient but improves the detection performance.
Third, for each pyramid level, we only aggregate the features from two neighboring levels as we assume that features in a lower level may be helpless for a higher level.
We compare DPE and PAFPN on four basic detectors~\cite{tian2019fcos,zhang2020bridging,li2020generalized,feng2021tood}.
As observed in~\cref{cmp_pafpn}, our DPE achieves more performance improvements than PAFPN~\cite{liu2018path}.
DPE outperforms the baselines by about $0.5{\sim}0.8$ AP with negligible extra computational cost, but PAFPN only achieves marginal improvements (about $0.2$), indicating the strong generality of our DPE. 
However, as the context is not decoupled, the improvements are still limited.

\subsection{Comparisons with State-of-the-arts}
\label{cmp_sota}

\begin{figure*}[t]
\begin{center}
    \includegraphics[width=\linewidth]{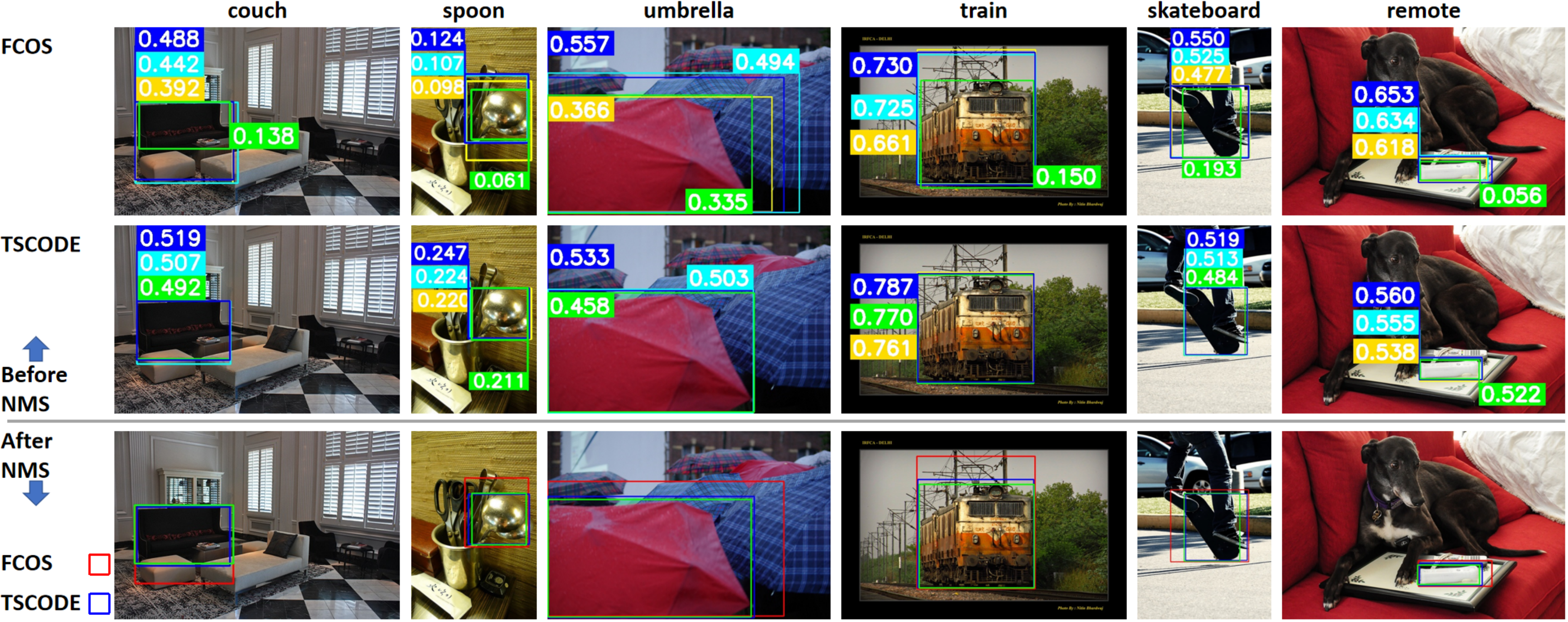}
\caption{Detection results before and after NMS predicted by FCOS~\cite{tian2019fcos} with and without~{\ours}. \textbf{TOP:}~Before NMS, the bounding boxes correspond to that in~\cref{fig:pic}. \textbf{Bottom:}~After NMS, the green boxes here mean ground-truth bounding boxes.}
\label{fig:vis}
\end{center}
\vspace{-0.6cm}
\end{figure*}

At last, we compare~\ours{} with recent state-of-the-art detectors on COCO~\textit{test-dev}. Here we select GFL~\cite{li2020generalized} as our baseline. Following common practice, we adopt $2{\times}$ schedule to train our models with standard multi-scale training strategy.
The results are reported with single-model single-scale testing for all methods.

As shown in~\cref{sota}, our method improves the performance of GFL to $46.7$ AP and $47.6$ AP with ResNet-101~\cite{he2016deep} and ResNeXt-101-32×4d~\cite{xie2017aggregated} backbones, respectively, outperforming all other methods~\cite{zhang2020bridging,kim2020probabilistic,ge2021ota,ma2021iqdet,li2021generalized}.
When using ResNeXt-101-64x4d~\cite{xie2017aggregated} backbone, the performance of~\ours{} can be further improved to $48.3$ AP. We also conduct more experiments by adopting Deformable Convolutional Networks (DCN)~\cite{dai2017deformable} to ResNeXt backbones. Following~\cite{zhang2021varifocalnet}, we replace the standard convolution in the last layer before prediction with deformable convolutions. Without bells and whistles, \ours{} achieves the AP of $50.8$ points, demonstrating the strong compatibility of our method with the advanced techniques in object detection.

\subsection{Qualitative Results}
In~\cref{fig:vis}, we qualitatively demonstrate how~\ours{} helps detectors improve their performance. For simplicity, we take FCOS~\cite{tian2019fcos} as an example. In the typical decoupled-head design, the classification branch and the localization branch share the same input features. However, their different preferences for context causes severe competition. As demonstated before, the feature context could lean to localization task, and thus detectors is prone to predict a poor classification confidence for the bounding box with high IoU. As a result, these high-quality bounding boxes is unlikely to survive after NMS.
This phenomenon is more significant on objects with texture-less surfaces, \eg, dark couch (the $1^{\text{st}}$ column) or shiny spoon (the $2^{\text{nd}}$ column).
Another example is the train (the $3^{\text{rd}}$ column), where more global context is required to successfully recognize the train. However, it is difficult to accurately localize the boundary of the train from a coarse feature map as the typical decoupled head does.
On the contrary, our \ours{} disentangles the feature context and generates feature inputs with specific feature context for each task. As a result, we can leverage more semantic context for classification and more detail and edge information for localization.

\begin{table}[tb]
\centering
\small
\begin{tabular}{l|lcc|c}
\toprule[1pt]
Method & AP & AP$_{50}$ & AP$_{75}$ & GFLOPs \\
\hline
\hline
FCOS~\cite{tian2019fcos} & 38.7 & 57.4 & 41.8 & 200.59 \\
FCOS w/ PAFPN & 38.7 & 57.4 & 41.7 & 206.49 \\
FCOS w/ DPE    & {\textbf{39.2}}$_{\textcolor{green}{+0.5}}$ & {\textbf{58.0}} & {\textbf{42.4}} & 213.62 \\
\hline
ATSS~\cite{zhang2020bridging} & 39.4 & 57.6 & 42.8 & 205.30 \\
ATSS w/ PAFPN & 39.6 & 58.3 & 42.2 & 211.20 \\
ATSS w/ DPE   & {\textbf{40.2}}$_{\textcolor{green}{+0.8}}$ & {\textbf{58.2}} & {\textbf{42.3}} & 217.89 \\
\hline
GFL~\cite{li2020generalized} & 40.2 & 58.4 & 43.3 & 208.39 \\
GFL w/ PAFPN & 40.4 & 58.6 & 43.6 & 214.29 \\
GFL w/ DPE   & {\textbf{40.8}}$_{\textcolor{green}{+0.6}}$ & {\textbf{58.8}} & {\textbf{43.9}} & 220.99 \\
\hline
TOOD~\cite{feng2021tood} & 42.4 & 59.8 & 46.1 & - \\
TOOD w/ PAFPN & 42.6 & 59.9 & 46.2 & - \\
TOOD w/ DPE & {\textbf{43.0}}$_{\textcolor{green}{+0.6}}$ & {\textbf{60.3}} & {\textbf{46.4}} & - \\

\bottomrule[1pt]
\end{tabular}
\caption{Comparison of our DPE and PAFPN~\cite{liu2018path}. FLOPs are measured on the input image size of $1280 \times 800$. Since TOOD~\cite{feng2021tood} contains complex deformable sampling~\cite{dai2017deformable}, we don't report its computational cost.}
\label{cmp_pafpn}
\vspace{-3mm}
\end{table}


\section{Conclusion}

In this paper, we have delved into the root causes of the inherent competition between classification and localization tasks and proposed a novel \ours{} to eliminate this competition. It decouples the semantic context of the two tasks through two efficient designs, SCE and DPE, and brings the features with richer semantic information for classification and with more edge information for localization. Extensive experiments on the MS COCO benchmark demonstrate the effectiveness the strong generality of~\ours{}.

{\small
\bibliographystyle{ieee_fullname}
\bibliography{tscode}
}

\end{document}